\documentclass{article}
\usepackage{graphicx} 
\usepackage{hyperref} 
\usepackage{enumitem} 

\title{The Cognitive Type Project - Mapping Typography to Cognition}
\author{Nik Bear Brown \\ \texttt{ni.brown@neu.edu} \\ 
\textsuperscript{1}Abecedarian, LLC \\
\textsuperscript{2}Northeastern University}
\date{March 2024}

\begin{document}

\maketitle

\section{The Cognitive Type Project}

\subsection{Abstract}

The Cognitive Type Project is focused on developing computational tools to enable the design of typefaces with varying cognitive properties. This initiative aims to empower typographers to craft fonts that enhance click-through rates for online ads, improve reading levels in children's books, enable dyslexics to create personalized type, or provide insights into customer reactions to textual content in media. A significant challenge in research related to mapping typography to cognition is the creation of thousands of typefaces with minor variations, a process that is both labor-intensive and requires the expertise of skilled typographers. Cognitive science research highlights that the design and form of letters, along with the text's overall layout, are crucial in determining the ease of reading and other cognitive properties of type such as perceived beauty and memorability. These factors affect not only the legibility and clarity of information presentation but also the likability of a typeface.

\medskip

Our research is committed to generating publicly available datasets and establishing foundational models that link the detailed anatomy of type with eye-tracking data from individuals interacting with text. By enriching existing datasets with insights into the physical and cognitive impacts of typefaces, we strive to illuminate the role of typography in reading comprehension and aesthetic appreciation. To this end, we are taking several approaches for easily creating cognitive type, that is, type that can be assessed for cognition. We have developed a lexical database mapping thousands of typographic terms to representational images. We use languages like Metafont and tools like Variable Fonts to create cognitively relevant glyphs. We use generative models to generate type and understand the typographic latent space.

\medskip

Finally, we are constructing a foundational model, inspired by AI systems like Midjourney and DALL·E, to facilitate the creation of an open-source text-to-type model. This model will enable typographers or researchers to specify the visual characteristics of a font, such as serif type, x-height, or bowl shape, translating the complex terminology of typeface classification into clear images for integration into typography software like FontForge or use in cognitive studies.

\subsection{Introduction }
Reading serves as a crucial mechanism for information acquisition and learning. The structure of letters and the overall design of typography play significant roles in the legibility of text, the clarity of information presentation, and the fluency of reading experiences. Research highlights the influence of typography on aspects such as legibility, comprehension, and aesthetic appeal (Beier et al., 2013; Beier et al., 2017; Bessemans, 2016a; Bessemans, 2016b; Bigelow, 2019; Brath and Banissi, 2016; Dressler, 2019; French et al., 2013; Gasser et al., 2005; Gasser et al., 2005; Kanfer and Ackerman, 1989; Larson et al., 2006; Larson and Picard, 2005; Lewis and Walker, 1989; Oppenheimer and Frank, 2008; Price et al., 2016; Pušnik et al., 2016; Wilkins et al., 2009; Woods et al., 2005). These studies indicate that font types not only affect the ease of reading but also contribute significantly to the retention and processing of information. Serif fonts, for example, have been shown to facilitate better recall than sans serif fonts, suggesting a profound impact of font choice on readability and comprehension. Despite the apparent arbitrariness in selecting fonts, it is clear that different typefaces yield distinct cognitive outcomes, with some enhancing readability and aesthetic appeal more than others. 

\medskip

However, the specific visual attributes of typefaces, such as serif styles or x-heights, and their direct effects on readability and aesthetic quality, have not been thoroughly investigated, highlighting the need for further research into how typography can enhance the reading experience and information retention. A deeper understanding of how font types affect recall and comprehension is essential for effectively conveying critical information. Studies have indicated that serif fonts tend to facilitate better recall of information than sans serif fonts, suggesting the profound impact font choice can have on readability and comprehension. While the selection of fonts may seem arbitrary, it's clear that different typefaces yield distinct cognitive outcomes, with certain ones enhancing readability and aesthetic appeal more significantly. Despite this, the specific visual attributes of typefaces, such as serif styles or x-heights, and their direct effects on readability and aesthetic quality, have not been thoroughly investigated. This gap in research underscores the need for further exploration into how typography can optimize the reading experience and information retention.

\subsection{Assessing the Cognitive Properties of Text}

Assessing the cognitive properties of text involves a variety of established techniques (Krafka K, et al., 2016; Dalmaijer, et al., 2014), each designed to measure how textual characteristics influence comprehension, recall, and engagement. These techniques include:

\begin{itemize}
    \item \textbf{Eye Tracking:} Measures where and for how long a reader looks at different parts of a text, providing insights into reading patterns, comprehension difficulties, and interests (Tobii Pro, 2017).
    \item \textbf{Reading Speed Tests:} Evaluate how quickly text can be read while maintaining comprehension. This can help in understanding the legibility and readability of different fonts or layouts.
    \item \textbf{Recall and Comprehension Tests:} After reading, participants are asked to recall information or answer questions about the text. This assesses how well information is understood and retained.
    \item \textbf{Dual-Task Methodology:} Involves having participants perform a secondary task while reading to measure cognitive load. The impact of text layout or typography on cognitive effort can be evaluated by how it affects performance on the secondary task.
    \item \textbf{fMRI and EEG:} Neuroimaging techniques like functional Magnetic Resonance Imaging (fMRI) and Electroencephalography (EEG) can observe brain activity in response to reading text. These methods can uncover the neural correlates of language processing and cognitive engagement.
    \item \textbf{Think-Aloud Protocols:} Participants verbalize their thoughts while reading, offering insights into their cognitive processes, strategies, and areas of difficulty.
    \item \textbf{Usability Testing:} In the context of digital texts, usability tests can assess how easily users can navigate, find information, and fulfill tasks, highlighting the cognitive impact of design choices.
    \item \textbf{A/B Testing:} Comparing two versions of a text to see which performs better in terms of reader engagement, comprehension, or preference. This can be particularly useful in digital environments for optimizing content presentation.
\end{itemize}

\begin{figure}[h]
\centering
\includegraphics[width=0.5\textwidth]{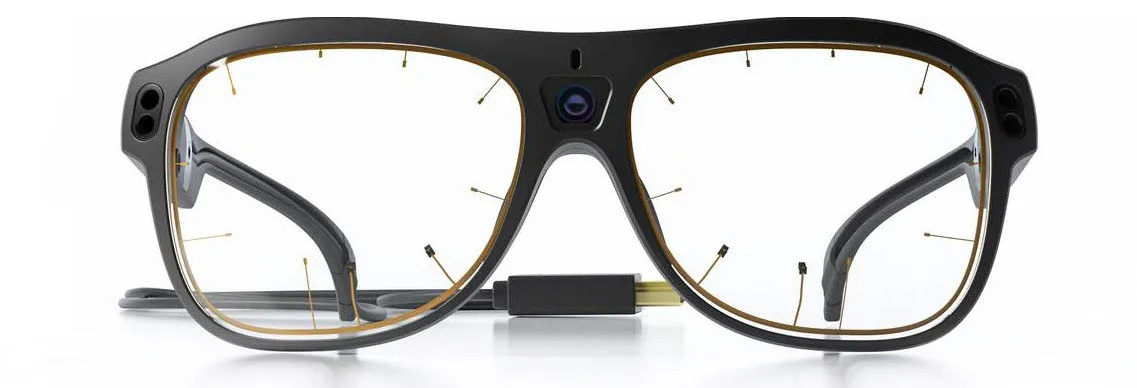}
\caption{Tobii Pro Glasses - Tobii Pro, 2017}
\label{fig:tobiipro}
\end{figure}

These methods can be used individually or in combination to provide a comprehensive understanding of how different aspects of text affect cognitive processing, engagement, and overall reading experience.

\subsection{Difficulties in Assessing the Cognitive Properties of Typefaces}

It is widely acknowledged that typefaces impact cognitive processes. However, the development of new typefaces is notoriously labor-intensive. The Abecedarian Classification of Typefaces ((Brown, N., 2024b) outlines a multitude of dimensions influencing typeface style. To determine which dimensions influence cognition, researchers require a method to efficiently produce characters with particular traits. While text-to-image models such as Midjourney and DALL·E have yielded impressive visuals, they lack training in the nuances of typography and tend to produce generic characters rather than typefaces with specific features. Creating figures like those in in the book Fonts and Encodings (Haralambous, Y., 2007) would be challenging using general-purpose text-to-image models like Midjourney, DALL·E, or Bing Image Creator, as they do not specialize in typographic nuances.

\medskip

\begin{figure}[h]
\centering
\includegraphics[width=0.9\textwidth]{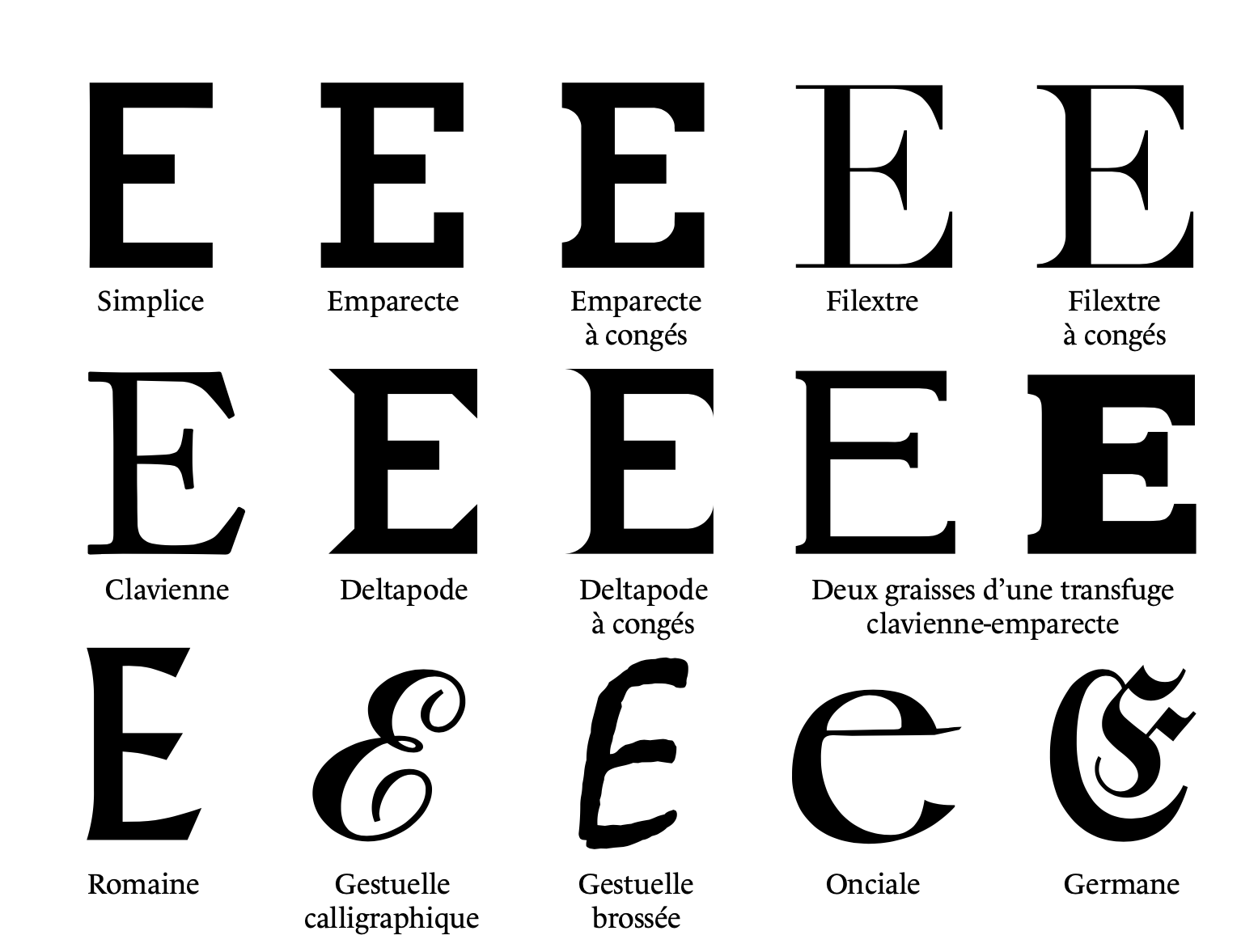}
\caption{Various typographic properties. Alessandrini’s dénominations préliminaires}
\label{fig:Alessandrini’s dénominations}
\end{figure}

\begin{figure}[h]
\centering
\includegraphics[width=0.9\textwidth]{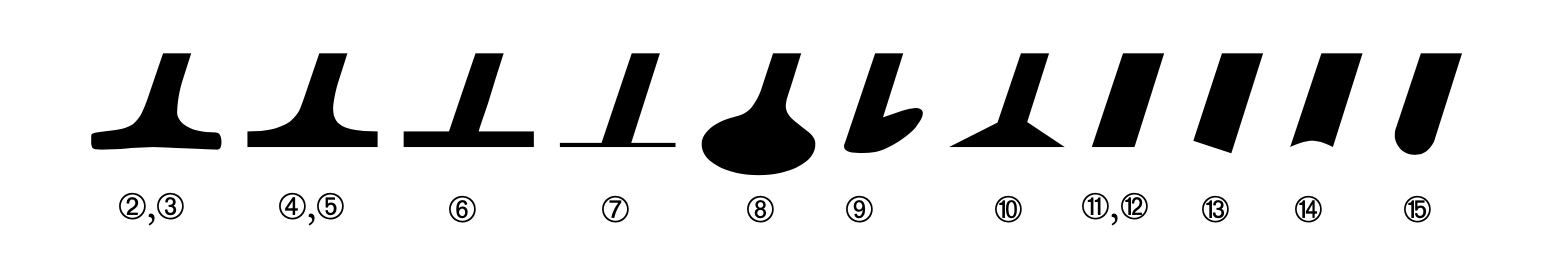}
\caption{Eleven types of serifs}
\label{fig:Eleven types of serifs}
\end{figure}

\begin{figure}[h]
\centering
\includegraphics[width=0.9\textwidth]{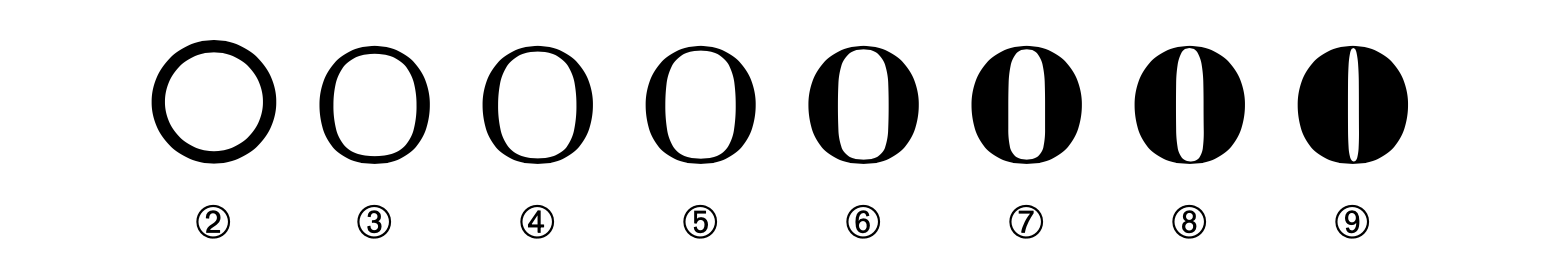}
\caption{The letter O from low to high contrast.}
\label{fig: letter O from low to high contrast.}
\end{figure}

\begin{figure}[h]
\centering
\includegraphics[width=0.9\textwidth]{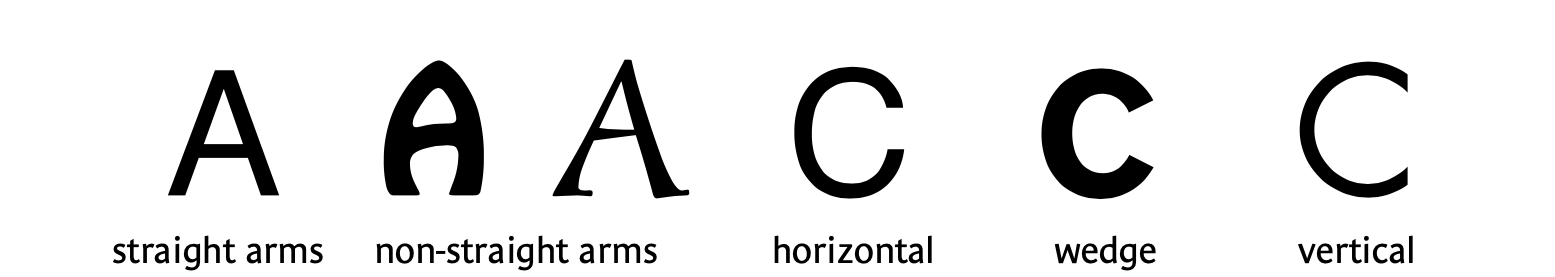}
\caption{Some arm and termination styles.}
\label{fig:Some arm and termination styles.}
\end{figure}

\begin{figure}[h]
\centering
\includegraphics[width=0.9\textwidth]{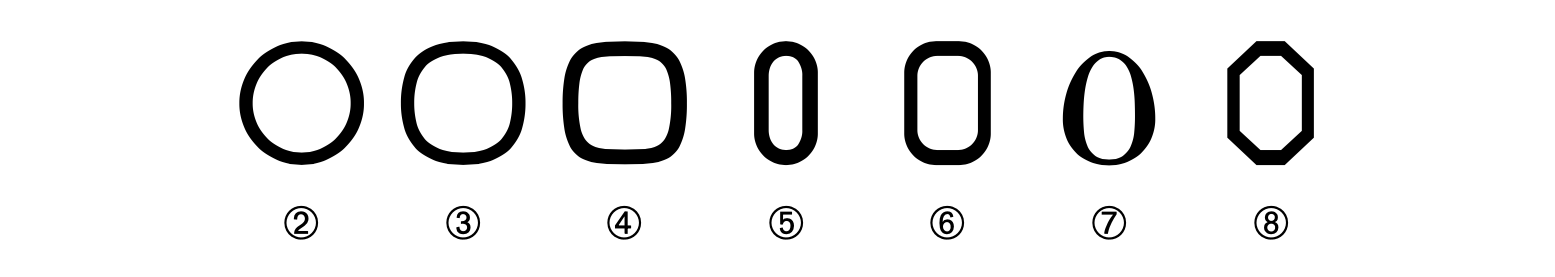}
\caption{Letter 'O' shapes (perfect circle, flattened circle, rounded square, square, etc.
Figures 2 through 6 are from (Haralambous, Y., 2007))}
\label{fig:Letter 'O' shapes}
\end{figure}

\medskip

\subsection{Creating Datasets Suitable for Assessing the Cognitive Properties of Typefaces}

In the quest to create datasets suitable for assessing the cognitive properties of typefaces, the Cognitive Type Project has embarked on an exploratory journey utilizing a variety of innovative tools and methodologies. The project has delved into the realms of Programmatic Typography, Typography-Specific Programming Languages, Hybrid Font Design Tools, Variable Fonts, and Generative Models and Deep Learning. Each of these avenues offers unique capabilities and insights into the intricate relationship between typography and cognition. The use of MetaFont (Metafont: Knuth, D. E., 2024), a program designed to define and generate bitmap fonts, has been instrumental in creating typefaces that can be finely tuned to explore cognitive impacts. This approach, rooted in mathematical descriptions of typeface anatomy, allows for an unprecedented level of precision in font design.

\medskip

Furthermore, the project has leveraged Hybrid Font Design Tools like Glyphs, FontLab (FontLab: FontLab Ltd., 2024), and RoboFont (RoboFont: Van Rossum, F., 2024), which blend graphical user interfaces with scripting capabilities, enabling the creation and modification of typefaces with a high degree of control and creativity. The advent of Variable Fonts has introduced a new dimension of flexibility, allowing for the dynamic adjustment of font characteristics such as weight, width, and slant through a single font file. This capability is vital for creating versatile datasets that can simulate a wide range of typographic conditions. Additionally, the integration of Generative Models and Deep Learning has opened up possibilities for identifying and generating novel typeface attributes that could influence cognitive processing. These technological approaches are paving the way for the development of "Text to Type" Foundational Models, which aim to transform the complex terminology of typeface classification into tangible, manipulable entities. By harnessing the strengths of MetaFont, Hybrid Font Design Tools, Variable Fonts, and Generative Models, the Cognitive Type Project is laying the groundwork for a new era of typography research, where the cognitive implications of type design are understood and utilized to their fullest potential. Below we overview the pros and cons of these approaches.

\subsubsection{Programatic Typography}

The realm of programmatic typography offers an intriguing avenue for the creation of typographic art, leveraging the power of code to draw and design. Tools such as Processing (Processing: Reas, C., et al., 2024), p5.js (P5.js: McCarthy, L., et al., 2024), OpenFrameworks (OpenFrameworks: The openFrameworks Team, 2024), NodeBox (NodeBox: Lieven van Velthoven, et al., 2024), and Cinder (Cinder: The Cinder Project Team, 2024) enable designers and programmers alike to experiment with and prototype unique typographic forms and patterns. Processing, for instance, is an accessible platform that introduces beginners to the creation of glyphs through simple coding principles, emphasizing experimentation over precision. Similarly, p5.js facilitates the creation of web-based and interactive typographic elements, making it an excellent tool for integrating typography with web technologies. 

\medskip

However, the creation of typefaces suitable for cognitive type—that is, typefaces designed with the understanding of how typographic form affects cognition—presents a more complex challenge. These tools, while wonderful for artistic endeavors, require a substantial amount of programming work to produce typefaces that are both aesthetically pleasing and functionally effective for cognitive purposes. OpenFrameworks and Cinder, with their extensive graphics libraries and capabilities for high-performance design, offer powerful resources for the creation of intricate glyph shapes. Yet, their steep learning curves and the necessity for experienced programming skills may pose barriers to those focusing solely on typography. NodeBox, with its Python-based platform geared towards generative design, excels in creating complex forms and is similarly positioned more towards typographic art than practical typeface development for cognitive applications. In essence, while these tools open up vast possibilities for artistic expression within typographic design, bridging the gap between artistic experimentation and the practical creation of typefaces optimized for cognitive enhancement remains a significant endeavor, necessitating a deep integration of design principles, cognitive science, and programming expertise.

\medskip

\begin{itemize}
    \item \textbf{Processing:} Enables beginners to create glyphs using simple coding principles, focusing on experimentation and prototyping. Processing is better suited to typographic art than cognitive type.
    \item \textbf{p5.js:} Allows for the creation of web-based and interactive typographic elements using JavaScript. It integrates easily with web technologies but is not well-suited for creating complete, production-ready fonts.
    \item \textbf{OpenFrameworks:} A C++ toolkit for high-performance typographic design. Offers a broad graphics library but has a steeper learning curve and is not focused on typography.
    \item \textbf{NodeBox:} Geared towards generative design, excellent for crafting complex typographic forms and patterns. It is Python-based and open-source but more suited to typographic art than cognitive type.
    \item \textbf{Cinder:} A C++ library for creating intricate glyph shapes but is not specifically geared towards typography. Requires experienced C++ programmers.
\end{itemize}

\subsubsection{Typography-Specific Programming Languages}

Metafont is a description language used to define raster fonts. It is also the name of the interpreter that executes Metafont code, generating bitmap fonts that can be embedded into, for example, PostScript. Metafont was devised by Donald Knuth as a companion to his TeX typesetting system.

\medskip

\begin{itemize}
    \item \textbf{Metafont:}
    \begin{itemize}
        \item \textbf{Purpose:} A language designed specifically for creating bitmap fonts.
        \item \textbf{Creator:} Devised by Donald Knuth as a complement to the TeX typesetting system.
        \item \textbf{Functionality:} Allows designers to define fonts programmatically with adjustable parameters and geometric equations.
        \item \textbf{Output:} Produces bitmap fonts, which are made up of pixels, making them resolution-dependent.
        \item \textbf{Specialization:} Uniquely tailored for typographic tasks, enabling the design of fonts through mathematical descriptions.
        \item \textbf{Integration:} Primarily used with TeX, providing a high degree of control over how characters are rendered in documents typeset with TeX.
    \end{itemize}
\end{itemize}

MetaPost (MetaPost: Hobby, J. D., 2024) refers to both a programming language and the interpreter of the MetaPost programming language. Both are derived from Donald Knuth's Metafont language and interpreter. MetaPost produces vector graphic diagrams from a geometric/algebraic description. The language shares Metafont's declarative syntax for manipulating lines, curves, points, and geometric transformations.

\medskip

\begin{itemize}
    \item \textbf{MetaPost:}
    \begin{itemize}
        \item \textbf{Purpose:} Based on Metafont, it focuses on creating precise technical illustrations and vector graphics.
        \item \textbf{Functionality:} Utilizes a similar syntax to Metafont but produces vector graphics, which are scalable and resolution-independent.
        \item \textbf{Output:} Generates diagrams and figures in PostScript, commonly used in technical and scientific documents.
        \item \textbf{Specialization:} Like Metafont, it is specialized for graphical tasks, particularly line drawings, which complements typographic designs.
        \item \textbf{Flexibility:} Can be used to draw shapes, plots, and various illustrations with mathematical precision, often used in academic and research settings.
    \end{itemize}
\end{itemize}

Both Metafont and MetaPost stand out in the realm of programming languages for their dedicated focus on typography and graphics, respectively. This specialization is rare among programming languages, which are more commonly designed for a broad range of computing tasks. Metafont and MetaPost offer a unique approach to design that is closely aligned with the mathematical precision and programmability required for high-quality typographic and illustrative work.

\medskip

However, to program in Metafont and MetaPost, an understanding of geometry, algebra, and Bézier curves is essential. Knowledge of these mathematical concepts enables users to craft detailed and sophisticated designs by specifying exact mathematical descriptions of the shapes. Metafont and MetaPost scripts often resemble mathematical formulas, which describe the paths and points that make up the characters and graphics.

\medskip

\begin{figure}[h]
\centering
\includegraphics[width=0.9\textwidth]{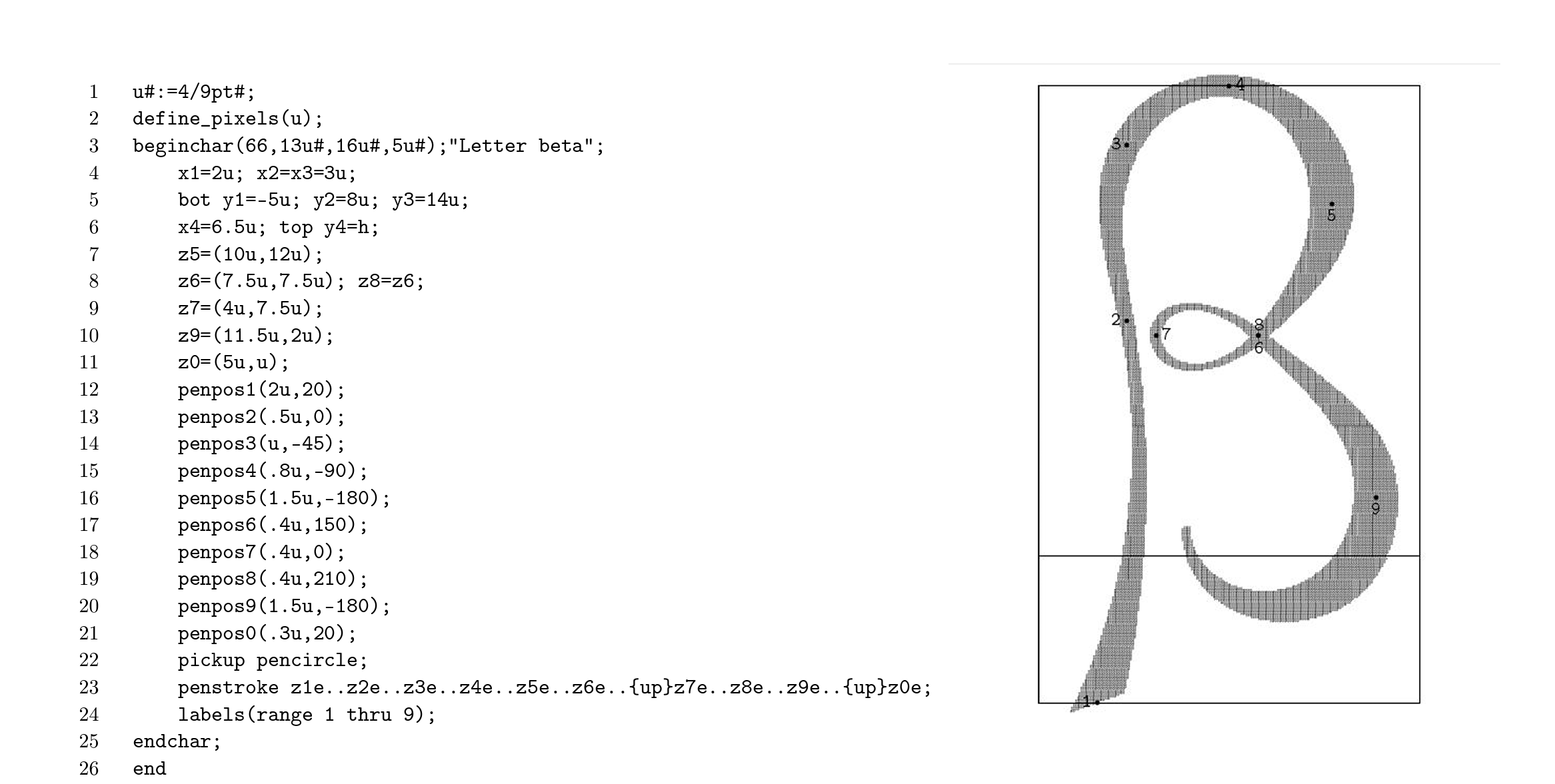}
\caption{The Greek Letter Beta ($\beta$) Programmed in MetaFont}
\label{fig:BetaProgrammedInMetaFont}
\end{figure}

\medskip

For this reason, the Cognitive Type project uses Metafont and MetaPost to create typographic training sets for generative models but feels it is unlikely to be adopted by typographers or cognitive scientists. Despite the collaboration of well-known type designers like Hermann Zapf with Knuth to create new fonts using Metafont, the system has not been widely adopted by professional type designers since its creation in 1982. Knuth attributes this to the complexity of requiring an artist to become proficient in mathematics to write a font with 60 parameters. Jonathan Hoefler commented that the Metafont system ultimately became "a technology behind zero of your favorite fonts." The Cognitive Type project is currently raising money to hire a Metafont and MetaPost programmer that can programmatically create millions of glyphs based on the Abecedarian Classification of Typefaces (Brown, N., 2024) for the training of "text to type" foundation models.

\subsubsection{Hybrid Font Design Tools}

Web and UI-based systems like Glyphs (Glyphs: Schneider, G., et al., 2024), FontLab (FontLab: FontLab Ltd., 2024), and RoboFont (RoboFont: Van Rossum, F., 2024) offer a graphical user interface for designing typefaces and allow for some level of scripting to extend functionality. These are powerful tools for making changes to typefaces, creating typefaces, and producing small datasets for generative models but still require learning the tools and making changes glyph by glyph.

\medskip

\textbf{Metapolator:}
\begin{itemize}
    \item \textbf{Description:} An open web tool aimed at streamlining the process of creating multiple fonts. It introduces an innovative approach by enabling work within a font design space, allowing designers to manage and manipulate many fonts simultaneously rather than focusing on individual glyphs or faces.
    \item \textbf{Features:} Supports a broad, project-level view of font design, facilitating rapid experimentation and development of font families.
    \item \textbf{Platform:} Web-based, accessible from any platform with internet access.
    \item \textbf{Scripting:} While primarily UI-driven, the open nature of Metapolator suggests potential for customization and extension by users familiar with web technologies.
    \item \textbf{Licensing:} The project and its fonts are under the GNU General Public License v3.0 (GPL), encouraging use and extension of the source code.
\end{itemize}

\textbf{Metaflop:}
\begin{itemize}
    \item \textbf{Description:} An accessible web tool for creating and modifying typefaces based on Metafont principles. It allows users to adjust font parameters through a user-friendly interface, generating unique typefaces.
    \item \textbf{Features:} Provides a modulator interface where users can tweak various aspects of a font's appearance through sliders and controls, effectively applying Metafont's parametric design principles.
    \item \textbf{Platform:} Web-based, ensuring wide accessibility without the need for specific operating system compatibility.
    \item \textbf{Scripting:} Leverages Metafont for backend processing, with the UI serving to abstract complex coding tasks into intuitive visual adjustments.
    \item \textbf{Licensing:} Source code and generated fonts are licensed under the GNU General Public License v3.0 (GPL), promoting open use and community contributions.
\end{itemize}

\begin{figure}[h]
\centering
\includegraphics[width=0.9\textwidth]{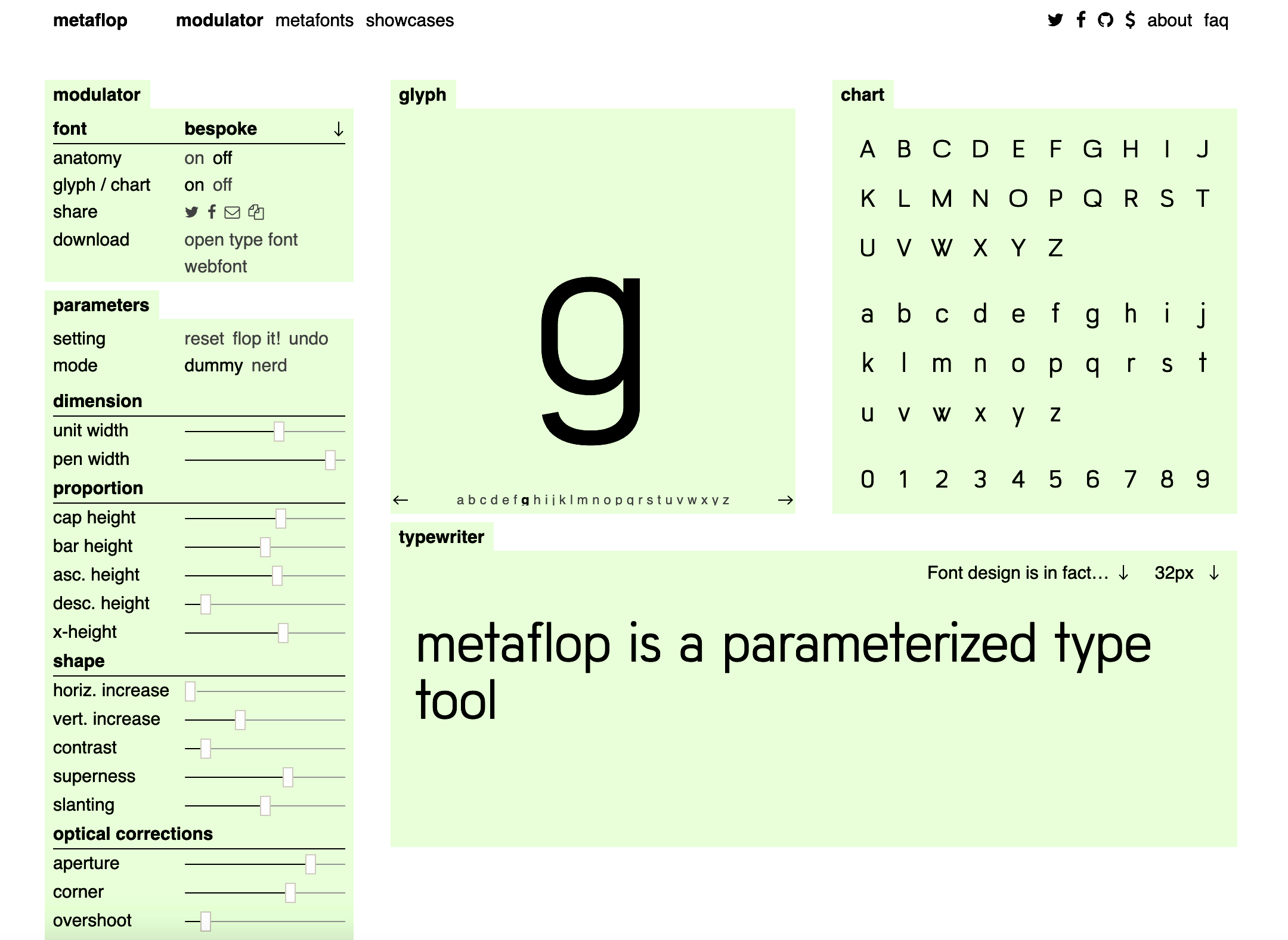}
\caption{Hybrid Font Design Tool - Metaflop}
\label{fig:Hybrid Font Design Tool - Metaflop}
\end{figure}

\textbf{Glyphs:}
\begin{itemize}
    \item \textbf{Description:} User-friendly software with a robust set of features for font design, scriptable with Python.
    \item \textbf{Platform:} macOS-specific, requiring purchase.
    \item \textbf{Features:} Offers a good balance for both beginners and professionals with its intuitive design and scripting capabilities.
\end{itemize}

\textbf{FontLab:}
\begin{itemize}
    \item \textbf{Description:} Professional software providing an extensive toolkit for crafting and refining typefaces.
    \item \textbf{Platform:} Proprietary, with a cost associated with its use.
    \item \textbf{Features:} Generates production-ready fonts and allows Python scripting for advanced customization.
\end{itemize}

\textbf{RoboFont:}
\begin{itemize}
    \item \textbf{Description:} Provides a highly customizable interface for designing glyphs, with Python scripting for extensive functionality.
    \item \textbf{Platform:} macOS exclusive, designed to fit into a modular design workflow.
    \item \textbf{Features:} Its clean UI and scripting capabilities make it versatile, though it may present a learning curve for non-coders.
\end{itemize}

These tools collectively represent a spectrum of options for typeface design, from web-based applications that democratize the design process to professional-grade software offering deep customization and precision. Each has its unique strengths, catering to different needs within the typographic community.

\subsubsection{Variable Fonts}

Variable fonts (Variable Fonts, 2024), also known as OpenType Font Variations, represent a significant advancement in font technology by allowing the customization of a font's appearance along multiple axes of variation. This flexibility means that instead of being restricted to a set number of pre-designed font weights, widths, and styles, users can fine-tune a font's characteristics to meet their specific needs. This capability not only enhances creative freedom but also optimizes efficiency, particularly in web typography, by consolidating multiple font variations into a single file, thus reducing overall file size.

\medskip

\textbf{Common Axes of Variation in Variable Fonts:}
\begin{itemize}
    \item \textbf{Weight (wght):} Controls the thickness of the strokes, ranging from thin to black. This axis allows for fine-tuning between light and bold appearances.
    \item \textbf{Width (wdth):} Adjusts the overall width of the font characters, from condensed to expanded, affecting the text's occupancy on a page or screen.
    \item \textbf{Slant (slnt):} Modifies the angle of the font, simulating italic styles without needing a separate italic font file. This axis tilts the letters to the right but differs from true italics in that it doesn't change the letterforms' design.
    \item \textbf{Italic (ital):} Enables a switch between upright and italic styles. Unlike slant, this axis can trigger a change to true italic letterforms if the font supports it.
    \item \textbf{Optical Size (opsz):} Adjusts the font's appearance for different text sizes, optimizing legibility across a range of sizes by altering character spacing, weight, and other details.
    \item \textbf{X-Height (xhgt):} Influences the height of lowercase letters relative to the font's overall size, affecting legibility and the visual density of text.
\end{itemize}

\medskip

These axes can be combined within a single variable font file, offering unprecedented control over typography with the potential for additional custom axes defined by type designers. The introduction of variable fonts thus marks a trans formative shift in the way fonts are used and managed, particularly in digital contexts where flexibility and efficiency are paramount.

\medskip

Variable fonts or OpenType Font Variations offer an innovative approach to typography by allowing designers and developers to adjust font characteristics dynamically through CSS. Here's how the primary axes of variation in variable fonts are manipulated using CSS:

\begin{itemize}
    \item \textbf{Weight (wght)}
    \begin{itemize}
        \item \textbf{CSS Attribute:} \texttt{font-weight}
        \item \textbf{Description:} Controls the thickness of the font strokes, offering a continuous range from light to bold. Instead of limited options like "normal" or "bold," any value within the font's weight range can be specified.
    \end{itemize}
\end{itemize}

\begin{figure}[h]
\centering
\includegraphics[width=0.9\textwidth]{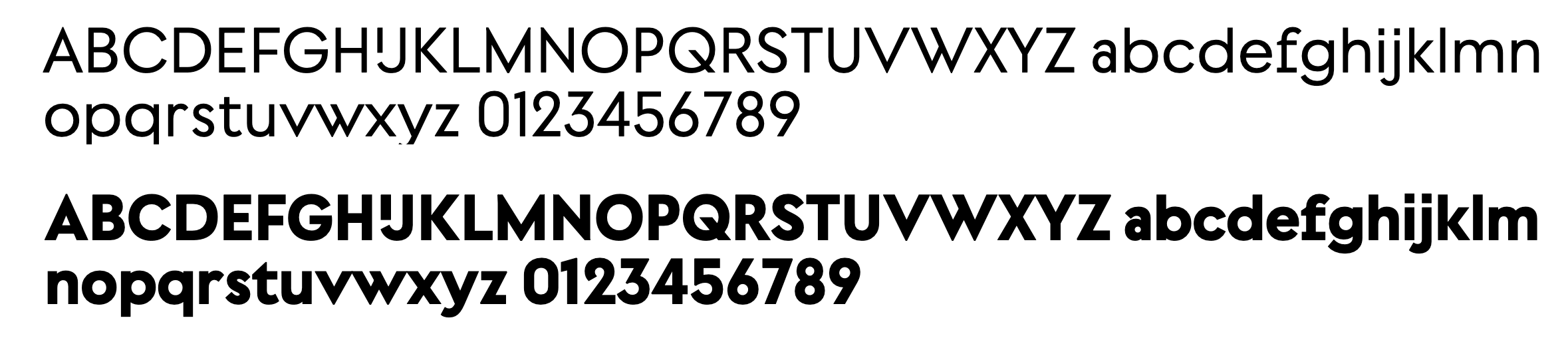}
\caption{Variable Fonts Axes of Variation - Weight (wght)}
\label{fig:Variable Fonts Axes of Variation - Weight (wght)}
\end{figure}

\begin{itemize}
    \item \textbf{Slant (slnt)}
    \begin{itemize}
        \item \textbf{CSS Attribute:} \texttt{font-style} for oblique styles, \texttt{font-variation-settings} for specific slant angles.
        \item \textbf{Description:} Provides a degree of slant to the font, without changing to italic letterforms. Useful for a subtle emphasis or stylistic choice.
    \end{itemize}
\end{itemize}

\begin{figure}[h]
\centering
\includegraphics[width=0.9\textwidth]{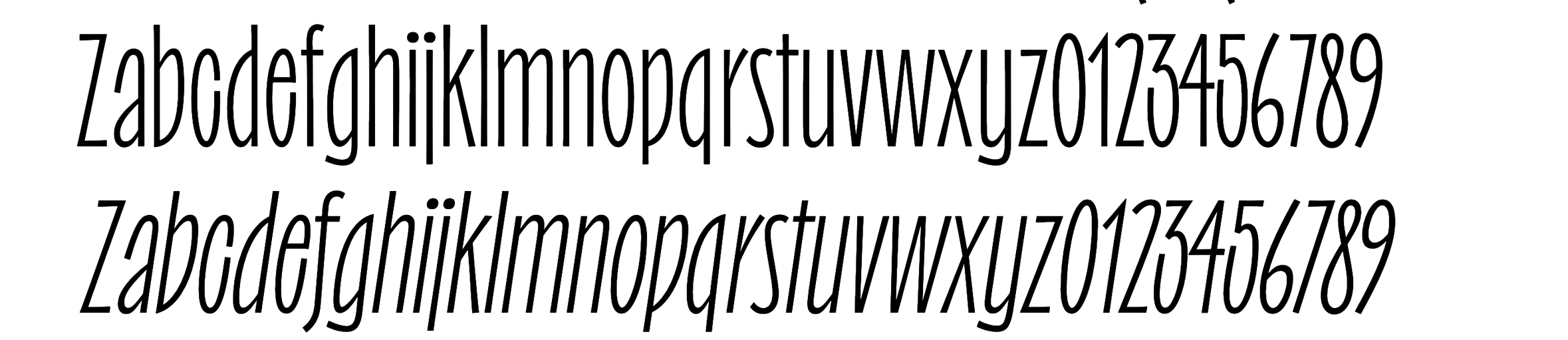}
\caption{Variable Fonts Axes of Variation - Slant (slnt)}
\label{fig:Variable Fonts Axes of Variation - Slant (slnt)}
\end{figure}

\medskip

Variable fonts are a powerful movement in typography, suited for creating font families through control of existing weight and slant parameters like (wght, wdth, slnt, ital, opsz). Adjusting the existing standard axes like weight (wght), width (wdth), slant (slnt), etc., can be done through parameters in CSS by anyone with basic knowledge of web development and design. These predefined axes are part of the OpenType specification and are included by the font designer within the variable font file.

\medskip

Defining a custom axis involves not only programming knowledge but also a deep understanding of typography and typeface design. It is a more advanced task that usually requires an experienced typographer or typeface designer. Creating a custom axis involves defining what the axis will control, designing the typeface variations that correspond to different values along the axis, and correctly implementing these variations within the font file itself. This process involves using font creation software such as Glyphs, FontLab, or RoboFont, which allows the designer to draw, interpolate, and test custom variations, extending beyond simple parameter adjustments into the realm of creative and technical typeface design.

\subsubsection{Generative Models and Deep Learning}

Neural Networks, particularly disentangled \(\beta\)-VAEs, can be used to identify latent features representing typeface letterforms, allowing for the potential discovery of new attributes for type classification (Issak et al., 2023). Our work in this area emphasizes that mapping the learned features of a model can reverse current heuristics and provide typographers with a new perspective on font classification. However, these models are sensitive to model structure, and much more needs to be done in this area to use this approach to extend existing type classification systems.

\medskip

Our work using simple generative models showed that one can create a dataset of glyphs similar to those in Google Fonts (Magre, N., \& Brown, N., 2022). This "TMINST" was composed of 565,292 MNIST-style grayscale images representing 1,812 unique glyphs across various styles of 1,355 Google fonts. While we can create many high-quality glyphs, we don’t have the prompting control of "text to image" foundational models like DALL·E and Stable Diffusion. Our belief is that while the approaches listed above are critical to training a "text to type" foundational model, the creation of such a model would allow anyone, including non-programmers and non-typographers, to create precise and detailed glyphs for the design of typefaces. In particular, to design "Cognitive Type" to be used by scientists to assess how differences in type affect cognition.

\subsection{"Text to Type" Foundational Models}

The development of a "Text to Type" foundational model is set to significantly impact typographic design, especially for the Cognitive Type project. This initiative strives to bridge the gap between the generative capabilities of current text-to-image models and the intricate demands of typography. Although present models can produce glyphs, they often lack the sophistication to comprehend typographic terms fully, which impedes the creation of "Cognitive Type" or professional font families with the necessary detail and consistency.

\medskip

Such a model would not only understand typographic terminology (see Figure 11) but also apply it according to type design principles. It would enable the creation of typefaces with specific attributes—such as weight, width, slant, and x-height—by interpreting descriptive text inputs. Examples include:

\medskip

\begin{itemize}
    \item Create a serif 'a' with a single-story structure and closed aperture.
    \item Design a slab serif 'g' with low stroke contrast and square terminals.
    \item Produce a humanist 'E' with open apertures and a double-story structure.
    \item Designing a lowercase 'a' with a slightly flared entry stroke and a pronounced curve at the top, suggestive of handwriting.
    \item Creating a Cyrillic character set that harmonizes visually with existing Latin characters, using diagonal stress and similar stroke weights.
    \item Designing a numeral '4' with a closed counter for better legibility at smaller sizes.
    \item Generating swash characters for uppercase 'A' and 'E' that add decorative touches while preserving legibility.
\end{itemize}

\begin{figure}[h]
\centering
\includegraphics[width=0.95\textwidth]{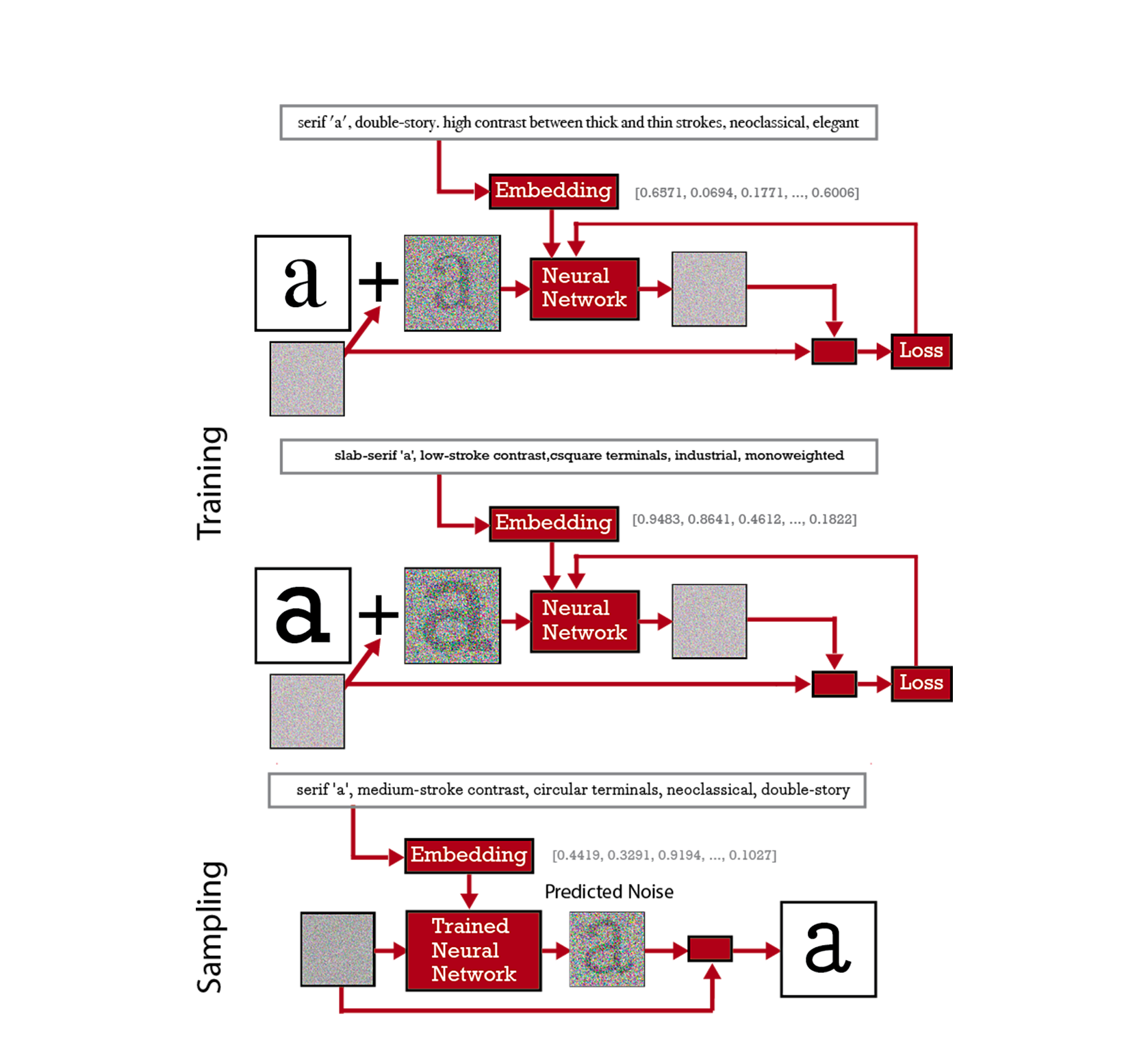}
\caption{Training a Neural Network to Understand Typographic Terms}
\label{fig:NN Typographic Terms}
\end{figure}

A "Text to Type" model could revolutionize typeface creation, making it more accessible to non-experts and opening new design possibilities. For the Cognitive Type project, it offers a chance to quickly prototype and test typefaces optimized for readability, legibility, and cognitive impact.

\medskip

The importance of a "Text to Type" model also extends to cognitive science, allowing for the systematic study of how typographic variations affect reading speed, comprehension, and aesthetic preference. This could lead to advances in visual cognition and the creation of typefaces tailored to specific audiences or conditions, such as dyslexia or low vision.

\medskip

Additionally, this model would address the labor-intensive nature of typeface creation and refinement. Automating parts of the design process would greatly reduce the time and expertise needed to develop new fonts, democratizing type design and allowing a wider range of creators to contribute. This aligns with the Cognitive Type project's goals to use technology to push the boundaries of typographic innovation and application.

\medskip

In conclusion, creating a "Text to Type" foundational model is crucial for the advancement of the Cognitive Type project. It is set to bring unprecedented efficiency, accessibility, and scientific rigor to typeface design, fostering new creative and research opportunities. By combining the powers of current generative technologies with the specific needs of typography, this model has the potential to spur a new era of innovation in the field.

\subsection{Summary}

The Cognitive Type Project delves into the development of computational tools to enable the creation of typefaces with varied cognitive properties. This ambitious initiative seeks to enhance typographers' ability to design fonts that not only improve user engagement and reading comprehension across various media but also cater to specific needs such as dyslexia-friendly typefaces. A crucial aspect of this research lies in the challenge of generating a vast array of typefaces with subtle differences, a task that demands both significant labor and the expertise of seasoned typographers. The project aims to bridge the gap in existing research by focusing on how the design and layout of text influence readability, aesthetic appeal, and memorability, factors that are essential for effective communication.

\medskip

To achieve its goals, the project is developing publicly accessible datasets and foundational models that integrate the detailed anatomy of type with eye-tracking data, thus offering new insights into the physical and cognitive impacts of typography. Utilizing tools like Metafont, Variable Fonts, and generative models, the team is exploring innovative ways to create cognitively relevant glyphs and understand the typographic latent space. A pivotal element of their approach involves constructing an open-source text-to-type model inspired by AI systems like Midjourney and DALL·E, which would allow for the specification of visual characteristics of fonts for use in cognitive studies and typography software.

\medskip

The Cognitive Type Project underscores the importance of typography in learning and information acquisition, highlighting the influence of font design on reading efficiency and comprehension. By employing techniques such as eye tracking, reading speed tests, and neuroimaging, researchers aim to comprehensively understand how textual characteristics impact cognitive processes. This multidisciplinary approach not only promises to enhance the design flexibility and accessibility of digital products but also supports the creation of "Cognitive Type" for scientific evaluation of typographic effects on cognition.

\medskip

However, the endeavor faces challenges, notably in the creation of new typefaces, as current text-to-image models lack the nuanced understanding of typography required for this task. The project advocates for the development of a "Text to Type" foundational model, which could revolutionize typeface design by making it accessible to a broader audience, including those without programming or typographic expertise. This model would facilitate the design of typefaces with precise cognitive objectives, contributing significantly to the fields of typography and cognitive science. In essence, the Cognitive Type Project is at the forefront of merging technological innovation with typographic design to explore how typography can optimize cognitive outcomes and information retention.

\subsection{References}

\begin{enumerate}[leftmargin=*,label=\arabic*.,itemsep=1ex,parsep=1ex]
    \item Beier, S., and Larson, K. (2013). How does typeface familiarity affect reading performance and reader preference? \textit{Inf. Design J.}, 20, 16--31. \href{https://doi.org/10.1075/idj.20.1.02bei}{doi: 10.1075/idj.20.1.02bei}
    \item Beier, S., Sand, K., and Starrfelt, R. (2017). Legibility implications of embellished display typefaces. \textit{Visible Lang.}, 51, 112--133.
    \item Bessemans, A. (2016a). Typefaces for children’s reading. \textit{TMG J. Media Hist.}, 19, 1--9. \href{https://doi.org/10.18146/2213-7653.2016.268}{doi: 10.18146/2213-7653.2016.268}
    \item Bessemans, A. (2016b). Matilda: a typeface for children with low vision. \textit{Digit. Fonts Reading 2016}, 8--34. \href{https://doi.org/10.1142/9789814759540_0002}{doi: 10.1142/9789814759540\_0002}
    \item Bigelow, C. (2019). Typeface features and legibility research. \textit{Vis. Res.}, 165, 162--172. \href{https://doi.org/10.1016/j.visres.2019.05.003}{doi: 10.1016/j.visres.2019.05.003}
    \item Brath, R., and Banissi, E. (2016). Using typography to expand the design space of data visualization. \textit{J Design Econ. Innov.}, 2, 59--87. \href{https://doi.org/10.1016/j.sheji.2016.05.003}{doi: 10.1016/j.sheji.2016.05.003}
    \item Brown, N. (2024a). The Cognitive Type Project - Mapping Typography to Cognition. Retrieved from \url{https://github.com/nikbearbrown/CognitiveType/tree/main/Papers/The_Cognitive_Type_Project_Mapping_Typography_to_Cognition}
    \item Brown, N. (2024b). The Abecedarian Classification of Type. Retrieved from \url{https://github.com/nikbearbrown/CognitiveType/tree/main/Papers/The_Abecedarian_Classification_of_Type}
    \item Brown, N. (2024c). The Abecedarian Visual Typographic Lexicon. Retrieved from \url{https://github.com/nikbearbrown/CognitiveType/tree/main/Papers/The_Abecedarian_Visual_Typographic_Lexicon}
    \item Brown, N. (2024d). Cognitive Type Project. Retrieved from \url{https://github.com/nikbearbrown/CognitiveType/} 
    \item Cinder: The Cinder Project Team. (2024) Cinder [Software]. Available from \url{https://libcinder.org/}
    \item Dalmaijer, E.S., Mathôt, S., \& Van der Stigchel, S. (2014). PyGaze: an open-source, cross-platform toolbox for minimal-effort programming of eye tracking experiments. \textit{Behavior Research Methods}, 46, 913-921. \href{https://doi.org/10.3758/s13428-013-0422-2}{doi:10.3758/s13428-013-0422-2}.
    \item Dressler, E. (2019). Understanding the Effect of Font Type on Reading Comprehension/Memory Under Time-Constrains. Omaha: University of Nebraska at Omaha.
    \item FontLab: FontLab Ltd. (2024) FontLab [Software]. Available from \url{https://www.fontlab.com/}
    \item French, M. M. J.; Blood, A.; Bright, N. D.; Futak, D.; Grohmann, M. J.; Hasthorpe, A.; Heritage, J.; Poland, R. L.; Reece, S.; \& Tabor, J. (2013). Changing fonts in education: How the benefits vary with ability and dyslexia. \textit{The Journal of Educational Research}, 106(4), 301-304. \href{https://doi.org/10.1080/00220671.2012.736430}{doi: 10.1080/00220671.2012.736430}.
    \item Gasser, M.; Boeke, J.; Haffernan, M.; Tan, R. (2005). The influence of font type on information recall. \textit{North American Journal of Psychology}, 7(2), 181-188.
    \item Gasser, M., Haffeman, J. B. M., and Tan, R. (2005). The influence of font type on information recall. \textit{N. Am. J. Psychol.}, 7, 181–188.
    \item Glyphs: Schneider, G., \& Seifert, E. (2024) Glyphs [Software]. Available from \url{https://glyphsapp.com/}
    \item Haralambous, Yannis. (2007). Fonts \& Encodings. O’Reilly Press. ASIN: 0596102429.
    \item Issak, A. A., Kakkar, S., Goetz, S., Brown, N. B., \& Harteveld, C. (2023, May). First TinyPapers track at The Eleventh International Conference on Learning Representations (ICLR). Kigali, Rwanda.
    \item Kanfer, R. \& Ackerman, P. L. (1989). Motivation and cognitive abilities: An integrative/aptitude-treatment interaction approach to skill acquisition. \textit{Journal of Applied Psychology}, 74, 657-690.
    \item Krafka K, Khosla A, Kellnhofer P, Kannan H, Bhandarkar S, Matusik W, \& Torralba A. (2016). Eye Tracking for Everyone. IEEE Conference on Computer Vision and Pattern Recognition (CVPR).
    \item Larson, K., and Picard, R. (2005). The Aesthetics of Reading. Available at: \url{https://affect.media.mit.edu/pdfs/05.larson-picard.pdf} (Accessed June 6, 2022).
    \item Larson, K., Hazlett, R. L., Chaparro, B. S., and Picard, R. W. (2006). Measuring the Aesthetics of Reading. Proceedings of HCI, People and Computers XX–Engage, 41–56.
    \item Lewis, C., and Walker, P. (1989). Typographic influences on reading. \textit{Br. J. Psychol.}, 80, 241–257. \href{https://doi.org/10.1111/j.2044-8295.1989.tb02317.x}{doi: 10.1111/j.2044-8295.1989.tb02317.x}
    \item Magre, N., \& Brown, N. (2022, February). Typography-MNIST (TMNIST): An MNIST-Style Image Dataset to Categorize Glyphs and Font-Styles. arXiv. Available from \url{http://arxiv.org/abs/2202.08112}
    \item McLean, R. (1997). The Manual of Typography. London: Thames and Hudson.
    \item Metafont: Knuth, D. E. (2024) Metafont [Software]. Stanford University. Available from \url{https://ctan.org/pkg/metafont?lang=en}
    \item Metaflop: Metaflop Project (2024) Metaflop [Software]. Available from \url{http://www.metaflop.com/}
    \item Metapolator: Metapolator Project. (2024) Metapolator [Software]. Available from \url{http://metapolator.com/}
    \item MetaPost: Hobby, J. D. (2024) MetaPost [Software]. Available from \url{https://tug.org/metapost.html}
    \item NodeBox: Lieven van Velthoven and Frederik De Bleser. (2024) NodeBox [Software]. Available from \url{https://www.nodebox.net/}
    \item OpenFrameworks: The openFrameworks Team. (2024) openFrameworks [Software]. Available from \url{https://openframeworks.cc/}
    \item Oppenheimer, D. M., and Frank, M. C. (2008). A rose in any other font would not smell as sweet: effects of perceptual fluency on categorization. \textit{Cognition}, 106, 1178–1194. \href{https://doi.org/10.1016/j.cognition.2007.05.010}{doi: 10.1016/j.cognition.2007.05.010}
    \item P5.js: McCarthy, L., et al. (2024) p5.js [Software]. Processing Foundation. Available from \url{https://p5js.org/}
    \item Price, J., McElroy, K., and Martin, N. J. (2016). The role of font size and font style in younger and older adults predicted and actual recall performance. \textit{Aging Neuropsychol. Cogn.}, 23, 366–388. \href{https://doi.org/10.1080/13825585.2015.1102194}{doi: 10.1080/13825585.2015.1102194}
    \item Processing: Reas, C., \& Fry, B. (2024) Processing [Software]. Processing Foundation. Available from \url{https://processing.org/}
    \item Pušnik, N., Podlesek, A., and Možina, K. (2016). Typeface comparison -- does the x-height of lower-case letters increased to the size of upper-case letters speed up recognition? \textit{Int. J. Ind. Ergon.}, 54, 164--169.
    \item RoboFont: Van Rossum, F. (2024) RoboFont [Software]. UFO Tools. Available from \url{https://robofont.com/}
    \item Tobii Pro (2017). Tobii Studio User’s Manual (Version 3.4.8). Stockholm: Tobii AB.
    \item Variable Fonts (as a technology): OpenType 1.8. (2024) Variable Fonts [Technology]. OpenType Specification. Available from \url{https://learn.microsoft.com/en-us/typography/opentype/otspec180/}
    \item Wilkins, A., Cleave, R., Grayson, N., and Wilson, L. (2009). Typography for children may be inappropriately designed. \textit{J. Res. Read.}, 32, 402–412. \href{https://doi.org/10.1111/j.1467-9817.2009.01402.x}{doi: 10.1111/j.1467-9817.2009.01402.x}
    \item Woods, R. J., Davis, K., and Scharff, L. F. V. (2005). Effects of typeface and font size on legibility for children. \textit{Am. J. Psychol. Res.}, 1, 86–102.
    
\end{enumerate}

\end{document}